\documentclass[10pt,twocolumn]{article}
\usepackage[margin=0.72in,columnsep=0.22in]{geometry}
\usepackage{booktabs}
\usepackage{multirow}
\usepackage{microtype}
\usepackage{graphicx}
\usepackage{xcolor}
\usepackage{amsmath,amssymb}
\usepackage{enumitem}
\usepackage{url}
\usepackage{xspace}
\usepackage[hidelinks]{hyperref}
\usepackage{tikz}
\usetikzlibrary{arrows.meta,positioning,fit}
\usepackage{pgfplots}
\usepgfplotslibrary{groupplots}
\pgfplotsset{compat=1.18}
\usepackage{pgfplotstable}
\usepackage{balance}
\hypersetup{
  pdftitle={SemanticFold: Latent Sequence Compression Separates Language Modeling, Decodability, and Reasoning},
  pdfauthor={Mingyan Liu; Min Huang},
  pdfsubject={Latent sequence compression and differential functional preservation}
}

\definecolor{sfblue}{HTML}{2563A6}
\definecolor{sforange}{HTML}{D97706}
\definecolor{sfgreen}{HTML}{27825B}
\definecolor{sfgray}{HTML}{64748B}

\newcommand{\Reff}{R_{\mathrm{eff}}}

\title{\textbf{SemanticFold: Latent Sequence Compression Separates\\
Language Modeling, Decodability, and Reasoning}}
\author{%
\begin{tabular}{cc}
Mingyan Liu & Min Huang\\
\texttt{msfocus@gmail.com} & \texttt{minhuang1@link.cuhk.edu.cn}
\end{tabular}}
\date{}

\begin{document}
\maketitle
\begin{abstract}
SemanticFold replaces selected adjacent hidden states with fewer latent states
inside a frozen decoder-only Transformer while leaving external tokenization
unchanged. We use this controlled intervention to ask what it means for model
capability to be preserved. Fixed-target language-model fit,
output-distribution proximity, task-relevant linear accessibility, and
reasoning behavior give different answers. In Qwen3-1.7B at $R=1.7$,
reasoning accuracy fell by 5.2 percentage points (95\% CI $[-9.8,-0.6]$),
while identical-target continuation NLL decreased by 0.135. An intervention
decomposition attributes that favorable NLL mainly to the learned residual
transform rather than shortening. On a 450-prompt Qwen population, tail
protection reduced mean
$D_{\mathrm{KL}}(p_{\mathrm{native}}\Vert p_{\mathrm{arm}})$ from 1.404 to
0.527, while the accuracy point estimate changed by $+1.56$ percentage points
versus the Full SemanticFold arm (95\% CI $[-3.11,6.22]$). Matched-$R$ controls on Qwen3-1.7B and
SmolLM2-1.7B found
no reliable cosine-boundary advantage over uniformly random legal boundaries
at $R\in\{1.3,1.7\}$. Matched checkpoint comparisons further show that
preservation profiles vary across model families and endpoints. Recovery on
one proxy metric therefore does not certify preservation on a different
functional endpoint; the evidence does not identify a universal compression
threshold or a complete causal mechanism.
\end{abstract}

\section{Introduction}
\label{sec:introduction}

If a Transformer processes fewer internal states while its tokenizer,
backbone weights, and output head remain fixed, what does it mean for its
capability to be preserved? A compressed model can retain teacher-forced
next-token fit while changing its free-running trajectory, expose an answer
to a linear decoder without using it behaviorally, or move its output
distribution toward the native model without recovering task accuracy.
Treating any one of these measurements as a certificate for the others hides
the scientific question.

SemanticFold makes that question experimentally accessible. Frozen lower
decoder layers produce \(N\) token-aligned contextual states. At layer \(K\),
an exact-budget policy selects non-overlapping adjacent pairs; a learned
compressor replaces every selected pair by one latent state, and frozen upper
layers process the resulting \(M<N\) states. External token IDs and
\emph{input-token granularity} do not change. Only the deeper
\emph{computational granularity} changes. The \(R=1\) path bypasses selection
and compression, providing a native-equivalence gate.

\begin{figure}[t]
\centering
\resizebox{\columnwidth}{!}{%
\begin{tikzpicture}[font=\scriptsize, node distance=2.5mm,
  box/.style={draw,rounded corners,minimum height=5mm,align=center},
  arr/.style={-{Latex[length=1.5mm]},thick}]
\node[box,fill=sfblue!10] (tok) {input\\\(N\) tokens};
\node[box,fill=sfblue!10,right=of tok] (low) {frozen lower\\layers};
\node[box,fill=sforange!15,right=of low] (fold) {exact-budget\\latent fold};
\node[box,fill=sfblue!10,right=of fold] (up) {frozen upper\\layers};
\node[box,fill=sfgreen!12,right=of up] (out) {logits\\\(M\) states};
\draw[arr] (tok)--(low);
\draw[arr] (low)--node[above] {\(h_{1:N}\)}(fold);
\draw[arr] (fold)--node[above] {\(\tilde h_{1:M}\)}(up);
\draw[arr] (up)--(out);
\node[below=4mm of fold,align=center,text width=4.6cm] (budget)
{\(F=N-M\) folds; input tokens unchanged; upper-layer sequence shortened};
\draw[arr,dashed] (budget)--(fold);
\end{tikzpicture}
}
\caption{SemanticFold changes deeper computational granularity without
changing tokenizer or API input length.}
\label{fig:architecture}
\end{figure}
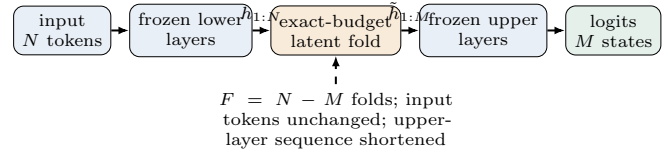

We evaluate four preservation questions. \emph{Fixed-target fit} asks whether
the same continuation tokens remain probable. \emph{Distributional
proximity} compares the complete next-token distribution with native
execution. \emph{Representational accessibility} asks whether task-relevant
labels remain linearly decodable from a specified hidden state.
\emph{Behavioral preservation} evaluates the model's own task decision or
generated trajectory. These endpoints are related but are not interchangeable
by definition.

The evidence spans frozen Qwen3, SmolLM2, and Pythia checkpoints. It includes
identical-target continuation scoring, paired reasoning populations,
cross-fitted answer probes, bounded generation, matched small-to-large
comparisons, intervention decompositions, and direct runtime measurements.
The central finding is a pattern of bounded dissociations. Qwen3-1.7B can
show lower fixed-target NLL while losing reasoning accuracy; the favorable
NLL is largely associated with the compressor's residual transform rather
than shortening. Tail protection can move the output distribution markedly
toward native execution without a corresponding reliable accuracy recovery.
Conversely, matched-\(R\) controls do not establish a reliable advantage for
cosine-selected over uniformly random legal boundaries. Preservation
therefore has to be reported as an endpoint- and checkpoint-specific profile.

Our contributions are:
\begin{itemize}[leftmargin=*,nosep,label=$\bullet$]
\item a controlled, exact-budget latent sequence-compression intervention in
frozen decoder-only Transformers, including explicit feasibility and rounding
semantics;
\item a multi-endpoint analysis separating fixed-target fit, distributional
proximity, task-relevant linear accessibility, behavior, and bounded
generation;
\item matched controls showing when proxy improvement fails to certify
behavioral recovery and when a boundary-selection advantage is not reliably
detected;
\item structural, checkpoint, and systems constraints that bound claims about
mechanism, scaling, memory reduction, and wall-clock speed.
\end{itemize}

\section{Related Work}
\label{sec:related}

\paragraph{Internal sequence shortening is prior art.}
Transformer computation has long been reduced by deleting, pruning, or
merging intermediate positions. PoWER-BERT progressively removes contextual
word vectors \cite{goyal2020powerbert}; DynamicViT and Token Merging develop
input-dependent pruning and similarity-based merging in vision
\cite{rao2021dynamicvit,bolya2023tome}. Closer language-model work includes
MrT5, whose learned gate deletes contextualized byte-level encoder states
\cite{kallini2025mrt5}; Dodo, which represents text with a dynamic number of
hidden states per decoder layer \cite{qin2024dodo}; LazyLLM, which prunes
prompt-token states during long-context inference \cite{fu2024lazyllm}; and
SlimInfer, which combines dynamic intermediate hidden-state pruning with
KV-cache management \cite{long2026sliminfer}. These works establish both
internal sequence shortening and variable internal granularity before
SemanticFold. We do not claim either primitive as new.

\paragraph{Prompt, latent-context, and adaptive-granularity methods.}
LLMLingua shortens the discrete prompt before target-model execution
\cite{jiang2023llmlingua}, whereas AutoCompressors learn continuous summary
vectors for preceding text segments \cite{chevalier2023autocompressors}.
Gist Tokens compress prompts into reusable internal activations through a
restricted attention mask, with reported prompt compression up to
$26\times$ \cite{mu2023gist}. ICAE encodes context into compact memory slots
that condition a target language model, reporting $4\times$ context
compression with latency and GPU-memory benefits \cite{ge2024icae}. These
methods demonstrate that prompts and contexts can be represented by shorter
learned internal sequences; SemanticFold does not claim that latent or
internal compression itself is new.

Activation Beacon is a particularly relevant systems-oriented comparison. It
progressively compresses long contexts directly into beacon-token key--value
activations and reports $2\times$ inference acceleration and an $8\times$
reduction in KV-cache memory across its long-context evaluations
\cite{zhang2024activationbeacon}. These engineering results use different
models, training, contexts, and evaluation settings and are not
apples-to-apples with SemanticFold's operating ratios. Activation Beacon
primarily develops and evaluates an efficient activation-compression
mechanism; SemanticFold uses controlled latent compression as an intervention
to test whether fixed-target fit, task-relevant linear accessibility,
reasoning, and free-running behavior remain aligned. We claim neither
compression-rate nor production-speed superiority.

ByteFlow replaces fixed subword tokenization with a trained byte-level
hierarchy and adaptive coding-rate boundaries \cite{deng2026byteflow}.
SemanticFold instead leaves external token IDs unchanged and replaces selected
adjacent contextual states at one intermediate boundary of an otherwise
frozen decoder. This distinction identifies the intervention; it is not a
claim that separating input and deeper computational granularity is itself
new.

\paragraph{Compression evaluation beyond perplexity.}
Prior compression studies have shown that preserving perplexity or related
local fidelity measures does not necessarily preserve downstream behavior.
Jaiswal et al. introduce LLM-KICK and show that small perplexity changes under
pruning or quantization can conceal large differences across knowledge,
reasoning, generation, retrieval, and summarization tasks
\cite{jaiswal-etal-2024-compressing}. Dutta et al. show that aggregate accuracy
can conceal instance-level answer flips and propose flips and KL divergence as
complementary distance measures \cite{dutta-etal-2024-accuracy}.
Xu et al. evaluate pruning and quantization across language-modeling,
downstream, and safety dimensions and argue for evaluation beyond perplexity
\cite{xu-etal-2024-beyond-perplexity}. Kurz et al. show that target-language
calibration can preserve perplexity and dominant representation structure
without consistently preserving downstream task performance
\cite{kurz-etal-2026-limitations}. Multi-metric evaluation of compressed
language models is therefore established prior work, not a contribution of
SemanticFold. Our narrower contribution is to apply a jointly controlled set
of fixed-target likelihood, output-distribution proximity, task-relevant
linear accessibility, behavior, and bounded-generation endpoints to one
frozen-backbone latent-sequence intervention, then test whether recovery of
one endpoint certifies another.

\paragraph{KV management, evaluation, and our positioning.}
H2O evicts selected KV entries during generative inference
\cite{zhang2023h2o}, while CALM reduces depth through calibrated early exit
\cite{schuster2022calm}. SemanticFold shortens the sequence processed by all
later attention and feed-forward blocks, but its current systems implementation
is not claimed to be production state of the art. Its primary contribution is
the controlled evaluation combination: frozen backbones, exact budgets, a
true $R=1$ bypass, identical-target continuation, out-of-fold task-local
probes, reasoning and bounded generation endpoints, matched random boundaries,
matched small-to-large comparisons, and an independent second family.
Structural and linear probes establish accessibility, not causal use
\cite{hewitt2019structural,elazar2021amnesic}. We use this design to ask which
functional properties survive latent compression, rather than to claim the
broad compression primitive.

\section{SemanticFold}
\label{sec:method}

\subsection{Exact-budget latent folding}

Let a frozen decoder map $N$ input tokens to hidden states
$H=(h_1,\ldots,h_N)$ at insertion layer $K$. The target budget is
computed as
\begin{align}
F&=\operatorname{round}_{\mathrm{even}}\!\left(N-\frac{N}{R}\right),
\qquad M=N-F,\label{eq:budget}\\
R_{\mathrm{eff}}&=\frac{N}{M}.
\end{align}
where $\operatorname{round}_{\mathrm{even}}$ is Python's half-to-even
rounding. Computing $F$ directly is part of the implementation contract; at
exact half-integers it need not equal $N-\operatorname{round}(N/R)$.
Short sequences can have $F=0$, in which case the runtime takes the native
bypass and does not invoke the compressor.

Every width-two fold reduces sequence length by one. The selector therefore
chooses exactly $F$ non-overlapping edges on the path of adjacent states.
For the default policy, edge scores are adjacent cosine similarities and an
exact-cardinality dynamic program maximizes their sum. Selected pairs become
$\tilde h_j=C_\theta(h_i,h_{i+1})$. The reference compressor also applies its
learned residual transform to singleton groups, so unselected states are not
bitwise copies. Pair-only and MLP-only controls isolate singleton copying and
residual adaptation, respectively.

\subsection{Compressor architecture and training}
\label{sec:compressor-training}

The same width-two compressor module is shared across every group in a run.
For a legal group $G$ of one or two $d$-dimensional states, it computes
\begin{align}
 q_i &= w_s^\top\operatorname{RMSNorm}_s(h_i),\qquad
 \alpha_i=\frac{\exp(q_i)}{\sum_{j\in G}\exp(q_j)},\\
 z_G &= \sum_{i\in G}\alpha_i h_i,\\
 C_\theta(G) &= z_G+W_2\operatorname{SiLU}
 \left(W_1\operatorname{RMSNorm}_o(z_G)\right).
\end{align}
The group input is a stack of source states, not a concatenation. The score
projection and both MLP projections have no bias; the implementation uses no
dropout and no custom initialization beyond the PyTorch module defaults. Both
RMSNorms use the checkpoint epsilon. The hidden width is 1,024 and the output
dimension remains $d$. Thus the module has $3d+2d(1{,}024)$ trainable
parameters: 4,200,448 when $d=2{,}048$ and 8,400,896 when $d=4{,}096$. For a
singleton, the softmax weight is one and the map reduces to
$h+W_2\operatorname{SiLU}(W_1\operatorname{RMSNorm}_o(h))$.

Each checkpoint uses a separately trained compressor while all backbone
parameters remain frozen. Training teacher-forces WikiText-derived
continuations and combines continuation cross-entropy, a KL term to the
uncompressed frozen teacher, and final-hidden-state cosine loss. The
checkpoint-specific data splits, grouping policy, coefficients, optimization
schedule, seeds, candidate checkpoints, and validation-only selection rules
are reported in Appendix~\ref{app:compressor-training}. No downstream
reasoning, probe, generation, or systems endpoint is used to select a
compressor.

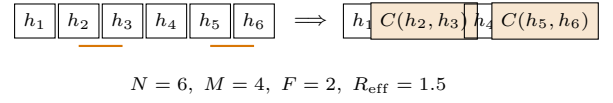
\begin{figure}[t]
\centering
\begin{tikzpicture}[font=\scriptsize,
  token/.style={draw,minimum width=5.2mm,minimum height=4.5mm},
  fold/.style={draw,fill=sforange!18,minimum width=9mm,minimum height=4.5mm}]
\foreach \x/\l in {0/1,0.58/2,1.16/3,1.74/4,2.32/5,2.90/6}
  \node[token] at (\x,0) {$h_{\l}$};
\draw[thick,sforange] (0.58,-0.33) -- (1.16,-0.33);
\draw[thick,sforange] (2.32,-0.33) -- (2.90,-0.33);
\node at (3.65,0) {$\Longrightarrow$};
\node[token] at (4.35,0) {$h_1$};
\node[fold] at (5.15,0) {$C(h_2,h_3)$};
\node[token] at (5.95,0) {$h_4$};
\node[fold] at (6.75,0) {$C(h_5,h_6)$};
\node[align=center] at (3.35,-0.85)
{$N=6,\;M=4,\;F=2,\;R_{\mathrm{eff}}=1.5$};
\end{tikzpicture}
\caption{Width-two folding on a path. Selected edges do not overlap, and the
dynamic program realizes the exact feasible fold count.}
\label{fig:fold}
\end{figure}

\subsection{Protection feasibility}

If $k$ positions must remain unfolded, the remaining $N-k$ positions can
support at most $\lfloor(N-k)/2\rfloor$ width-two folds. Hence a necessary
condition is
\begin{equation}
F\leq\left\lfloor\frac{N-k}{2}\right\rfloor,
\qquad k\leq k_{\max}=N-2F.
\label{eq:kmax}
\end{equation}
For a contiguous protected tail with all preceding edges legal, this bound is
also attainable. The quantity $k_{\max}$ is task-independent geometry.
Whether a selected fold is ``safe'' depends on which positions are declared
protected and can therefore depend on the endpoint. We report requested and
realized protection separately.

\subsection{Positions, caching, and equivalence}

A retained latent state carries a frozen original-coordinate position; it is
not interpreted as a newly tokenized string. Rotary position calculations use
those semantic coordinates. At generation step $t$, lower-layer caches have
length $N+t$, upper-layer caches have length $M+t$, and the new token has
semantic position $N+t$ in both regions.

\begin{figure}[t]
\centering
\begin{tikzpicture}[font=\scriptsize,
  bar/.style={draw,minimum height=5mm,anchor=west},
  arr/.style={-{Latex[length=1.5mm]},thick}]
\node at (0,0.45) {lower layers};
\node[bar,fill=sfblue!18,minimum width=5.4cm] at (1.0,0.45)
{$N$ prefill states $+\;t$};
\node at (0,-0.25) {upper layers};
\node[bar,fill=sforange!20,minimum width=3.7cm] at (1.0,-0.25)
{$M$ prefill states $+\;t$};
\draw[arr] (6.55,0.45) -- (6.95,0.45);
\draw[arr] (4.85,-0.25) -- (5.25,-0.25);
\node[align=left] at (4.85,-0.95)
{new token semantic position: $N+t$\\upper physical index: $M+t$};
\end{tikzpicture}
\caption{Upper layers store fewer prefill states, while generation appends one
state per step to every layer.}
\label{fig:cache}
\end{figure}
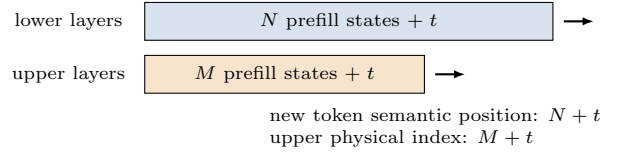

We distinguish bitwise identity, whole-trajectory exactness, and functional
equivalence. Bitwise identity requires same-shape tensors to match exactly.
Whole-trajectory exactness requires every generated token decision to match.
Functional endpoints use pre-specified paired accuracy, regression, recovery,
NLL, probe, or agreement criteria. Passing one contract does not imply
passing another.

\section{Experimental Design}
\label{sec:experiments}

\subsection{Frozen models and populations}

All evidence uses public decoder checkpoints, frozen model weights, and one
model-specific compressor per checkpoint. Execution uses BF16 on an RTX 4090;
the Qwen3-8B study uses no quantization or CPU offload. The scientific
populations are frozen independently, so rows from different populations are
not treated as repeated measurements of one common effect.

\begin{table*}[t]
\centering
\caption{Principal checkpoints and endpoint populations. Counts denote prompts
or records; a dash means that endpoint was not run.}
\label{tab:config}
\small
\begin{tabular}{lrrrrrrrr}
\toprule
Model & Params & Layers & $K$ & Fixed NLL & Reasoning & Probe & Generation & Systems\\
\midrule
Qwen3-1.7B & 1.7B & 28 & 6 & 50 & 500/450 & 300 & 100 & 1K--8K\\
SmolLM2-1.7B & 1.7B & 24 & 7 & 48 & 300/450 & 240 & 100 & 2K--8K\\
Qwen3-8B & 8.2B & 36 & 8 & 50 & 300 & 300 & 50 & 2K; 8K bounded\\
Pythia-1.4B & 1.4B & 24 & 5 & 50 & 300 & 300 & 20 & --\\
Pythia-6.9B & 6.9B & 32 & 7 & 50 & 300 & 300 & 20 & --\\
\bottomrule
\end{tabular}
\end{table*}

Every study has a true $R=1$ oracle in which selection and compression are
bypassed. The confirmatory Qwen and Pythia gates require bitwise-equal hidden
states and logits. Checkpoint revisions and population construction are
summarized in the appendix. Exact prompt texts, labels, token IDs, and
machine-readable evaluation records are retained in the accompanying
reproducibility artifacts.

\subsection{Endpoints answer different questions}

\paragraph{Fixed-target language-model fit.}
A frozen prefix is executed natively or compressed, and both arms
teacher-force the same continuation token IDs. Qwen3-1.7B uses 50 records with
192 prefix and 64 continuation tokens at $R=1.7$; SmolLM2 uses 48 records with
96 prefix and 32 continuation tokens at $R=1.2$. Qwen3-8B and both Pythia
checkpoints use separately frozen 50-record populations. The primary value is
compressed-minus-native mean NLL.

\paragraph{Behavior and linear accessibility.}
Reasoning accuracy is scored on frozen finite-label task populations. For the
450-prompt intervention population, nine BBH tasks contribute 50 prompts each:
date understanding, disambiguation QA, formal fallacies, logical deduction
with five and seven objects, penguins in a table, reasoning about colored
objects, temporal sequences, and tracking shuffled objects with five objects.
Prompts end in an answer cue; every candidate label must tokenize to exactly
one token after a leading space. Accuracy is the argmax over those candidate
logits at the final prompt position, and the margin is the correct-label logit
minus the largest competing-label logit. No chain-of-thought or free-form
generation is scored in this endpoint. The answer probe is an L2-regularized
multinomial logistic regression applied to
the normalized final prompt hidden state, with task-local classes,
five-fold task-stratified cross-fitting, train-fold-only standardization, and
pre-specified accuracy, macro-F1, and one-vs-rest macro-AUC. The probe tests
linear accessibility at one representation and does not establish causal use
by the backbone.

\paragraph{Output-distribution proximity.}
On the 450-prompt Qwen finite-label population, Native, Full SemanticFold,
Pair-only, MLP-only, and Tail protection arms are scored at the same final prompt
position. We report accuracy and
$D_{\mathrm{KL}}(p_{\mathrm{native}}\Vert p_{\mathrm{arm}})$. Prompt-level
bootstrap intervals for mean and median DKL use 10,000 paired draws; accuracy
uses the existing task-stratified paired procedure.

\paragraph{Generation and systems.}
Bounded greedy diagnostics record first-token and trajectory agreement,
normal output, immediate EOS, loops, printable-garble flags, and numerical
finiteness. Systems experiments separately measure structural KV reduction,
Native-versus-SemanticFold TTFT/E2E latency, and cache-path fidelity.

\subsection{Matched controls}

The fixed-continuation decomposition holds prefixes, continuations,
checkpoints, and selected boundaries fixed while comparing Native, Full
SemanticFold, Pair-only exact singleton copying, MLP-only residual adaptation
without folding, and a no-training mean-merge baseline.

The matched-boundary control uses the same first 300 finite-label frozen
prompts for Qwen3-1.7B and SmolLM2-1.7B at $R\in\{1.3,1.7\}$. For every prompt,
cosine and uniformly random legal boundaries share the exact $N$, $M$, $F$,
checkpoint, insertion layer, compressor, label readout, and generated work.
One deterministic random matching is derived per prompt from seed 20261005.
The primary contrast is paired cosine-minus-random accuracy with 10,000
prompt-level, task-stratified bootstrap draws.

Matched small-to-large comparisons use an internally identical 300-prompt
population within each family. Qwen3 compares 1.7B with 8B; Pythia compares
1.4B with 6.9B. Difference-in-differences estimates are descriptive of those
checkpoint pairs and do not isolate parameter count.

\subsection{Inference policy}

Intervals are percentile intervals from pre-specified paired bootstrap
procedures unless stated otherwise. Families are not pooled. A confidence
interval containing zero is not treated as evidence of equivalence.

\section{What Is Preserved Under Latent Compression?}
\label{sec:results}

\subsection{Endpoint profiles}

Table~\ref{tab:endpoint-profile} places the principal endpoint changes on one
page. The populations are frozen separately, so the table compares
preservation profiles rather than estimating a joint multivariate effect.

\begin{table*}[t]
\centering
\caption{Compression-associated endpoint changes at selected operating points.
Values are compressed minus Native with 95\% paired bootstrap intervals.}
\label{tab:endpoint-profile}
\small
\begin{tabular}{llrrr}
\toprule
Model & $R$ & Reasoning accuracy & Fixed-target NLL & Answer-probe macro AUC\\
\midrule
Qwen3-1.7B & 1.7 & $-.0520\;[-.0980,-.0060]$ & $-.1353\;[-.2156,-.0593]$ & $-.0268\;[-.0662,.0125]$\\
SmolLM2-1.7B & 1.2 & $-.0400\;[-.0767,-.0033]$ & $+.0127\;[-.0446,.0832]$ & $+.0224\;[-.0306,.0768]$\\
Qwen3-8B & 1.3 & $-.0233\;[-.0733,.0267]$ & $-.0852\;[-.1247,-.0492]$ & $-.0839\;[-.1133,-.0548]$\\
Pythia-1.4B & 1.3 & $-.0333\;[-.0867,.0200]$ & $+.0004\;[-.0143,.0153]$ & $+.0202\;[-.0152,.0556]$\\
Pythia-6.9B & 1.3 & $+.0067\;[-.0433,.0600]$ & $-.0037\;[-.0213,.0144]$ & $-.0316\;[-.0686,.0043]$\\
\bottomrule
\end{tabular}
\end{table*}

Qwen3-1.7B gives the clearest within-checkpoint dissociation: reasoning
accuracy declined while identical-target NLL decreased and the specified
answer-probe AUC had no detected decline. SmolLM2's fixed-target NLL was
approximately unchanged, yet its original reasoning
population also declined. The 450-prompt SmolLM2 intervention population had
the opposite point direction (Native 0.2244, Full SemanticFold 0.2667), showing
that behavioral direction depends on the tested population and cannot define a
universal threshold.

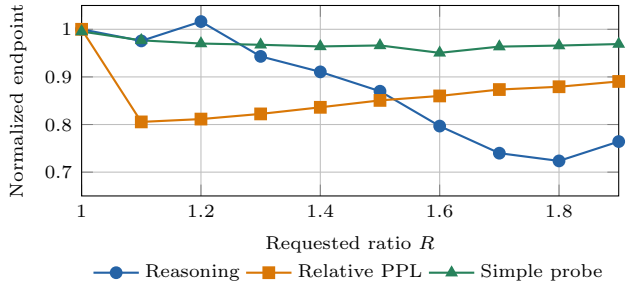
\begin{figure}[t]
\centering
\begin{tikzpicture}
\begin{axis}[
  width=\columnwidth,height=4.1cm,
  xlabel={Requested ratio $R$},
  xmin=1,xmax=1.9,xtick={1,1.2,1.4,1.6,1.8},
  ylabel={Normalized endpoint},
  ymin=0.65,ymax=1.05,
  legend style={font=\scriptsize,at={(0.5,-0.30)},anchor=north,
    legend columns=3,draw=none},
  tick label style={font=\scriptsize},
  label style={font=\scriptsize},
  grid=major]
\addplot[sfblue,thick,mark=*] table[
  x=requested_R,y expr=\thisrow{reasoning_accuracy}/0.41,
  col sep=comma]{figures/qwen_dissociation.csv};
\addlegendentry{Reasoning}
\addplot[sforange,thick,mark=square*] table[
  x=requested_R,y=relative_ppl,col sep=comma]
  {figures/qwen_dissociation.csv};
\addlegendentry{Relative PPL}
\addplot[sfgreen,thick,mark=triangle*] table[
  x=requested_R,y=probe_auc,col sep=comma]
  {figures/qwen_dissociation.csv};
\addlegendentry{Simple probe}
\end{axis}
\end{tikzpicture}
\caption{The original Qwen rate sweep normalized to $R=1$. Endpoints move
non-monotonically and do not share one compression threshold.}
\label{fig:qwen}
\end{figure}

\subsection{Fixed-target fit}

Teacher-forcing identical continuations rules out a target-selection
explanation for the NLL results. Qwen3-1.7B at $R=1.7$ changed by $-0.13529$;
SmolLM2 at $R=1.2$ changed by $+0.01266$. Qwen3-8B remained favorable at
$R=1.2$, 1.3, and 1.5 with changes $-0.09910$, $-0.08521$, and $-0.04164$.
Pythia NLL was effectively unchanged at both scales. These are
teacher-forced fit measurements, not claims of improved general language
capability.

\subsection{Task-relevant linear accessibility}

For Qwen3-1.7B at $R=1.7$, behavior changed by $-0.0467$ on the probe
population, while answer-probe accuracy changed by $-0.0100$ (95\% CI
$[-0.0600,0.0367]$) and macro AUC by $-0.0268$. For SmolLM2 at $R=1.2$,
behavior changed by $-0.0500$, probe accuracy by approximately zero, and AUC
by $+0.0224$. Among native-correct/compressed-incorrect prompts, the
compressed out-of-fold probe recovered the gold class for 22/50 Qwen prompts
and 13/22 SmolLM2 prompts. The representation can therefore retain linearly
accessible answer information when the model's own decision changes; this
does not show that the backbone uses that information. High probe separability
also does not establish that compression created a semantic code rather than a
task-correlated linear readout of the measured state.

Matched checkpoint comparisons further bound generalization. In Qwen3, the
small-to-large macro-AUC interaction was $-0.0796$ (95\% CI
$[-0.1207,-0.0381]$). Pythia reproduced the negative interaction direction,
$-0.0518$ (95\% CI $[-0.1010,-0.0028]$), but absolute Pythia probe
separability was weak: 15/100 and 94/100 permutation AUCs equaled or exceeded
the observed compressed AUC at 1.4B and 6.9B. These are checkpoint-pair
interactions, not a parameter-count scaling law.

\subsection{Bounded generation}

On 100 frozen SmolLM2 prompts at $R=1.3$, compressed generation produced 93
normal outputs, two immediate-EOS outputs, two token loops, and three
printable-garble flags. Cached and stateless paths agreed on all first tokens
and on 85/100 complete 16-token trajectories; neither produced NaN or Inf.
On 50 Qwen3-8B prompts at $R=1.3$, both Native and compressed runs were
mechanically normal, but first tokens agreed for 31/50 and full trajectories
for 9/50. Mechanical health therefore does not imply semantic or trajectory
equivalence. These diagnostic populations are bounded and do not estimate a
universal failure rate.

\section{When Preservation Metrics Disagree}
\label{sec:disagreement}

The central empirical result is a disagreement among preservation endpoints.
Fixed-continuation likelihood, output-distribution proximity, linear
accessibility, and task accuracy answer different questions. None is an
interchangeable certificate for the others.

\subsection{Where the favorable likelihood comes from}

Table~\ref{tab:nll-decomposition} separates sequence shortening from the
learned residual transform on identical continuation targets. MLP-only applies
the transform without shortening; Pair-only shortens paired states but copies
singletons; Mean merge shortens at the same boundaries without the learned
compressor.

\begin{table*}[t]
\centering
\caption{Fixed-continuation NLL decomposition. Parentheses give the paired
95\% confidence interval for the change from Native. Lower is better.}
\label{tab:nll-decomposition}
\small
\begin{tabular}{llcc}
\toprule
Model & Arm & NLL & $\Delta$NLL vs. Native (95\% CI)\\
\midrule
Qwen3-1.7B & Native & 2.6783 & ---\\
 & Full SemanticFold & 2.5391 & $-0.1391\;[-0.2218,-0.0625]$\\
 & Pair-only & 2.5766 & $-0.1017\;[-0.1780,-0.0329]$\\
 & MLP-only & 2.4575 & $-0.2207\;[-0.3017,-0.1475]$\\
 & Mean merge & 2.7795 & $+0.1012\;[+0.0763,+0.1280]$\\
\midrule
SmolLM2-1.7B & Native & 2.7231 & ---\\
 & Full SemanticFold & 2.7357 & $+0.0127\;[-0.0458,+0.0829]$\\
 & Pair-only & 2.7602 & $+0.0372\;[-0.0076,+0.0999]$\\
 & MLP-only & 2.6587 & $-0.0643\;[-0.0953,-0.0345]$\\
 & Mean merge & 2.9945 & $+0.2714\;[+0.0955,+0.5014]$\\
\bottomrule
\end{tabular}
\end{table*}

MLP-only has 0.0816 lower NLL than Full SemanticFold for Qwen
($[0.0591,0.1072]$) and 0.0770 for SmolLM2
($[0.0300,0.1439]$). Thus the favorable Qwen likelihood cannot be assigned
to shortening alone: residual adaptation supplies a substantial benefit, and
folding offsets part of it. Mean merge worsens NLL in both models, so arbitrary
shortening at the selected boundaries is also insufficient. Because MLP-only
changes every prefix state without shortening the sequence, this decomposition
identifies generic learned state adaptation as a contributor to NLL; it does
not identify sequence compression as the cause of the favorable Full-versus-
Native contrast.

\subsection{Distributional proximity does not certify behavior}

For each eligible Qwen prompt, we compute
$D_{\mathrm{KL}}(p_{\mathrm{Native}}\Vert p_{\mathrm{arm}})$ at the answer
position over the model vocabulary. Table~\ref{tab:dkl-behavior} uses the same
450 finite-label prompts at $R=1.7$. Accuracy intervals are task-stratified;
KL intervals use 10,000 prompt bootstrap resamples.

\begin{table*}[t]
\centering
\caption{Output-distribution proximity and finite-label behavior on Qwen3-1.7B.
Accuracy changes are relative to Full SemanticFold.}
\label{tab:dkl-behavior}
\scriptsize
\begin{tabular}{lcccc}
\toprule
Arm & Accuracy & $\Delta$ accuracy (95\% CI) & Mean $D_{\mathrm{KL}}$ (95\% CI) & Median $D_{\mathrm{KL}}$ (95\% CI)\\
\midrule
Native & .3444 & --- & .000 $[.000,.000]$ & .000 $[.000,.000]$\\
Full SemanticFold & .2978 & --- & 1.404 $[1.345,1.465]$ & 1.238 $[1.185,1.287]$\\
Pair-only & .2889 & $-.0089\;[-.0356,.0178]$ & 1.281 $[1.225,1.336]$ & 1.160 $[1.107,1.217]$\\
MLP-only & .3489 & $+.0511\;[.0000,.1044]$ & .097 $[.090,.105]$ & .062 $[.058,.067]$\\
Tail protection & .3133 & $+.0156\;[-.0311,.0622]$ & .527 $[.492,.562]$ & .497 $[.456,.539]$\\
\bottomrule
\end{tabular}
\end{table*}

Tail protection moves the answer distribution substantially toward Native,
but its accuracy change relative to Full SemanticFold remains uncertain. MLP-only moves
both endpoints in the favorable direction. The combined result is mixed:
distributional proximity contains useful information, but is an incomplete
predictor of behavioral recovery under these interventions.

The probe results create a related distinction. A linear decoder tests whether
specified answer information remains accessible at one representation, whereas
task accuracy tests what the frozen downstream computation actually produces.
Cross-fitting and finite permutation controls bound the probe claim, but do not
turn accessibility into causal use. Consequently, favorable NLL, low KL, high
cosine similarity, probe separability, and answer accuracy should be reported
as separate endpoint-specific results.

\section{Boundary Choice and Feasibility Geometry}
\label{sec:boundary-geometry}

Boundary policy can matter only within the legal fold budget. We therefore
separate two questions: whether cosine-selected boundaries outperform a
uniform legal matching at the same ratio, and whether a requested protection
rule is geometrically feasible.

\subsection{Matched-ratio boundary control}

Table~\ref{tab:matched-random} compares cosine selection with one deterministic
uniform legal matching on the same 300 finite-label prompts. Within every row,
the checkpoint, compressor, readout, targets, $N$, $M$, and $F$ are identical;
only the boundary policy changes. Intervals use 10,000 task-stratified paired
bootstrap resamples.

\begin{table*}[t]
\centering
\caption{Cosine versus random legal boundaries under matched compression ratios.}
\label{tab:matched-random}
\small
\resizebox{\textwidth}{!}{%
\begin{tabular}{lccccc}
\toprule
Model & $R$ & Native & Cosine & Random & Cosine $-$ Random (95\% CI)\\
\midrule
Qwen3-1.7B & 1.3 & .3633 & .3533 & .3333 & $+.0200\;[-.0367,.0767]$\\
Qwen3-1.7B & 1.7 & .3633 & .3167 & .2933 & $+.0233\;[-.0400,.0867]$\\
SmolLM2-1.7B & 1.3 & .2667 & .3133 & .2900 & $+.0233\;[-.0400,.0867]$\\
SmolLM2-1.7B & 1.7 & .2667 & .2533 & .2633 & $-.0100\;[-.0667,.0467]$\\
\bottomrule
\end{tabular}}
\end{table*}

Every interval crosses zero, and the point estimates do not show a consistent
ratio dependence. These data do not establish a reliable accuracy advantage
for cosine selection. The SmolLM2 $R=1.7$ row lies outside its validated operating range and is an exploratory matched control. Earlier comparisons
at Qwen $R=1.7$ and SmolLM2 $R=1.2$ could not isolate checkpoint effects from
ratio effects; the matched rows remove that ambiguity without identifying a
mechanism or an optimal policy.

\subsection{Protection feasibility}

The reference implementation computes
\begin{equation}
 F=\operatorname{round}_{\mathrm{even}}\!\left(N-\frac{N}{R}\right),
 \qquad M=N-F,
\end{equation}
where Python's half-to-even rule is applied before deriving $M$. This can differ
from rounding $N/R$ at exact half-integers. Unit checks at $R=2$ give
$F=2,4,4,6$ for $N=5,7,9,11$, respectively.

If $k$ positions are protected, the remaining $N-k$ positions can support at
most $\lfloor(N-k)/2\rfloor$ disjoint adjacent folds. Hence
\begin{equation}
 F\leq \left\lfloor\frac{N-k}{2}\right\rfloor,
 \qquad k_{\max}=N-2F.
\end{equation}
This is a geometric capacity bound. Actual fold safety also depends on which
positions are protected and on the legal-boundary definition used by the
selector.

\begin{table}[t]
\centering
\caption{Tail protection feasibility over the evaluated 450-prompt populations.}
\label{tab:feasibility}
\small
\resizebox{\columnwidth}{!}{%
\begin{tabular}{lccccc}
\toprule
Model & $R$ & $k_{\max}$ min/med/max & Requested & Realized & Safe\\
\midrule
Qwen3-1.7B & 1.7 & 7/27/48 & 16 & 7--16 & 1.00\\
SmolLM2-1.7B & 1.3 & 25/89/152 & 16 & 16 & 1.00\\
\bottomrule
\end{tabular}}
\end{table}

All 900 evaluated records satisfy the fold-safety check. Some Qwen prompts cannot
protect all 16 requested tail positions because their exact fold budgets leave
only seven to fifteen protected degrees of freedom. The bound explains when an
intervention can be realized; it is not a causal threshold for preservation or
a universal safe compression ratio.

\section{What the Evidence Says About Mechanism}
\label{sec:mechanisms}

The endpoint dissociation is better established than its cause. We evaluate
each mechanistic proposal as a scoped intervention: a negative result rejects
the tested operationalization, not every mechanism in the same family.

\begin{table*}[t]
\centering
\caption{Mechanism-directed evidence and its warranted scope.}
\label{tab:mechanisms}
\small
\begin{tabular}{p{0.31\textwidth}p{0.12\textwidth}p{0.47\textwidth}}
\toprule
Tested proposition & Decision & Evidence and scope\\
\midrule
Low similarity/high variance marks risky folds & Supported & Held-out diagnostic association under single-group restoration; no causal intervention or deployable policy.\\
Online control removes domain rate drift & Supported & Exact-budget control reduced requested-versus-realized rate error across the audited domains.\\
A learned gate gives stable held-out benefit & Unsupported & Eight-prompt test gains changed sign across compressor seeds.\\
Oracle-safe boundaries leave recovery headroom & Supported & Oracle accuracy reached 93.2\% at $R=1.5$; privileged labels preclude a deployable claim.\\
Oracle benefit reflects stable pair structure & Unsupported & Pair interactions had weak cross-seed rank and sign stability and did not reconstruct the oracle.\\
A single-fold adapter restores reasoning & Unsupported & Accuracy did not improve and perplexity worsened; multi-fold adaptation was not tested.\\
A compact localized patch restores reasoning & Unsupported & No eligible candidate passed the pre-specified selection criterion; the held-out test set was not evaluated.\\
The tested addressability intervention restores reasoning & Unsupported & Accuracy change $-0.0259$, 95\% CI $[-0.1037,0.0574]$ on 60 training/validation prompts.\\
Structural features add reproducible prediction & Unsupported & The pre-specified joint OOF-$R^2$/bootstrap-MAE criterion failed.\\
\bottomrule
\end{tabular}
\end{table*}

The addressability intervention produced strong structural contrast on every
tested prompt in the training and validation sets, yet did not improve the
reasoning endpoint. The oracle establishes recovery headroom, but learned gating, stable pairwise
structure, a single-fold adapter, and localized patches did not pass their
criteria. Together these results rule out several simple explanations without
locating the complete causal pathway.

The decomposition controls add two constraints. Copying singleton states
exactly did not reliably restore finite-label accuracy: relative to Full SemanticFold,
Pair-only changed Qwen accuracy by $-0.0089$ (95\% CI
$[-0.0356,0.0178]$) and SmolLM2 by $+0.0022$
($[-0.0244,0.0311]$). Protecting final prompt positions shifted Qwen's output
distribution toward Native, but the behavioral interval still crossed zero;
SmolLM2 did not show the same point-estimate direction. Thus neither singleton
rewriting nor bounded tail preservation alone explains the behavioral change.

Matched-ratio random boundaries further show that exact cardinality does not
determine the point estimate, while providing no reliable evidence that cosine
selection is superior. The native-to-compressed probe-transfer result remains
a representation-geometry observation rather than a demonstrated mechanism.
Distributed state replacement, position removal, later attention, and
checkpoint-specific representation use remain viable explanations.

\section{Systems Consequences}
\label{sec:systems}

SemanticFold creates three distinct systems claims: a structural reduction in
upper-layer state and KV length, fidelity of the cached implementation, and
measured wall-clock behavior. They require separate evidence.

\subsection{Cached execution preserves the tested function}

The cache study compares stateless recomputation with heterogeneous cached
SemanticFold, not compression with Native. Qwen's strongest common functional
band is \textsc{Strict}; SmolLM2 has zero accuracy difference and perfect answer
agreement at both tested ratios. Whole-trajectory exactness remains below one,
so the evidence supports fidelity under frozen criteria rather than universal
numerical identity.

\begin{table}[t]
\centering
\caption{Cached versus stateless SemanticFold.}
\label{tab:cache}
\small
\begin{tabular}{llrrrr}
\toprule
Model & $R$ & Stateless & Cached & Agree & Traj.\\
\midrule
Qwen & 1.1 & .417 & .417 & .997 & .967\\
Qwen & 1.2 & .400 & .403 & .990 & .957\\
Qwen & 1.3 & .430 & .433 & .990 & .970\\
SmolLM2 & 1.2 & .290 & .290 & 1.000 & .960\\
SmolLM2 & 1.3 & .280 & .280 & 1.000 & .970\\
\bottomrule
\end{tabular}
\end{table}

\subsection{Memory reduction and latency are separate}

After prefill, lower layers retain $N$ prompt states and upper layers retain
$M$. At decode step $t$, their cache lengths are $N+t$ and $M+t$. This geometry
reduces upper-layer KV memory by construction. Wall-clock latency additionally
includes boundary selection, compression, dispatch, hardware utilization, and
ordinary decode.

\begin{table}[t]
\centering
\caption{SemanticFold versus Native at $R=1.3$. Times are medians in
milliseconds; ratios are SemanticFold/Native.}
\label{tab:runtime}
\scriptsize
\resizebox{\columnwidth}{!}{%
\begin{tabular}{llrrrrr}
\toprule
Model & Context & TTFT N/SF & TTFT ratio & E2E N/SF & E2E ratio & KV save\\
\midrule
Qwen1.7 & 2K & 163.35/128.65 & .788 & 575.70/483.57 & .840 & 18.01\%\\
Qwen1.7 & 8K & 2024.55/1442.52 & .713 & 2516.78/1811.70 & .720 & 18.09\%\\
SmolLM2 & 2K & 61.03/61.74 & 1.012 & 260.57/273.28 & 1.049 & 16.24\%\\
SmolLM2 & 8K & 299.25/284.84 & .952 & 581.74/560.19 & .963 & 16.31\%\\
\bottomrule
\end{tabular}}
\end{table}

Qwen shows clear reductions in this local implementation. SmolLM2 has small
overhead at 2K and small reductions at 8K despite similar structural memory
savings. Each condition uses BF16 on one RTX 4090, exactly 16 generated tokens,
one warmup, and five retained alternating-order measurements. These are bounded
local timings, not a universal speedup or production comparison.

Additional cache geometry and measurement details appear in Appendix~\ref{app:systems-protocol}.

\section{Discussion}
\label{sec:discussion}

\subsection{Preservation is endpoint-specific}

SemanticFold separates several meanings of preservation that are often treated
as one. Fixed-target NLL evaluates teacher-forced fit to identical
continuations. Answer-position KL measures proximity of one output
distribution. A probe measures accessibility to a specified decoder. Accuracy
and free-running generation measure behavior of the complete frozen
computation. The experiments show that these quantities can move in different
directions.

The NLL decomposition makes this distinction concrete. Residual adaptation
without shortening is more favorable than Full SemanticFold in both 1.7B
models, while mean merging at the same boundaries is worse. The favorable Qwen
NLL is therefore neither evidence that shortening is harmless nor evidence
that any merge suffices. Likewise, Qwen tail protection reduces mean
native-to-intervention KL from 1.404 to 0.527 but does not yield a reliable
accuracy recovery. Better fit and closer output distributions remain useful
measurements with narrower meanings than behavioral preservation.

High global representation similarity does not resolve the disagreement.
Mean paired cosines exceed 0.95 in every matched model while task-local probe
behavior changes. Linear accessibility is also not causal use
\cite{elazar2021amnesic}. These observations motivate local geometric analysis,
but the negative gating, patching, pair-structure, and addressability results
leave the mechanism open.

\subsection{Checkpoint profiles differ without defining a scaling law}

The matched Qwen3 comparison shows a negative small-to-large macro-AUC
interaction, and the Pythia comparison reproduces its direction; both paired
bootstrap intervals exclude zero. Pythia's absolute probe separability is weak,
so the replication concerns the interaction direction rather than strong
absolute decoding. Accuracy does not show one common cross-family pattern.

These results establish checkpoint variation, not a universal size effect.
Each family contains only two checkpoints, and parameter count changes with
width, learned weights, training history, representation dimension, insertion
depth, and compressor fit. Baseline separability also differs between
families. The matched-ratio boundary control adds that cosine selection has no
reliable accuracy advantage over a uniform legal matching at the tested
ratios. It therefore cannot explain the checkpoint interactions.

\subsection{The useful systems claim is bounded}

Tokenization fixes the external input units but need not fix the number of
states processed by every deeper layer. SemanticFold provides an exact-budget
intervention in which $N$ contextual states become $M<N$ latent states before
later blocks. This design space is shared with prior work on internal pruning,
merging, compression, and adaptive segmentation
\cite{kallini2025mrt5,qin2024dodo,fu2024lazyllm,long2026sliminfer,deng2026byteflow}.

The structural consequence is shorter upper-layer state and KV sequences.
Realized latency remains implementation- and hardware-dependent: Qwen improves
in the local 2K and 8K measurements, whereas SmolLM2 has slight overhead at 2K
and a small improvement at 8K. These local measurements do not establish a
universal latency advantage or superiority to production serving systems.

\subsection{Validation must follow the intended use}

A deployment ratio should be selected for a particular checkpoint and endpoint,
not from NLL or compression rate alone. The exact feasibility bound determines
whether a protection rule can coexist with a requested fold budget, but does
not determine whether the resulting computation is safe. Likewise, cached-path
fidelity under frozen tests does not imply universal trajectory identity.

A stronger future study would vary ratio, insertion layer, compressor seed, and
boundary policy within each checkpoint; measure fixed-target, distributional,
representational, and behavioral endpoints on the same population; and use
interventions that distinguish information loss from downstream failure to use
retained information. The present evidence motivates that design without
claiming that a universal latent granularity already exists.

\section{Limitations}
\label{sec:limitations}

All conclusions are endpoint-scoped. Fixed-continuation NLL covers frozen
WikiText-derived targets; answer-position KL covers one position and direction
of divergence; probes are linear and task-local; finite-label accuracy and
short greedy generations cover bounded behaviors. None establishes unchanged
calibration, factuality, long-form coherence, general capability, or causal
use of a representation.

The model sample remains small. Matched small-to-large evidence comes from two
Qwen3 and two Pythia checkpoints, with architecture, weights, width, training,
insertion depth, representation dimension, and compressor fit entangled.
Parameter count is not isolated, and no universal scaling law follows.
Absolute Pythia probe separability is weak, while Qwen3-8B reasoning intervals
are underpowered for small effects.

The new matched-ratio boundary comparison covers two 1.7B checkpoints and only
$R=1.3$ and $R=1.7$. It uses one deterministic random legal matching per
prompt; all cosine-minus-random accuracy intervals cross zero. It neither
proves equivalence nor identifies an optimal boundary policy. The SmolLM2
$R=1.7$ condition is exploratory and outside its validated operating range.

The KL analysis is position-specific and asymmetric. A lower
$D_{\mathrm{KL}}(p_{\mathrm{Native}}\Vert p_{\mathrm{arm}})$ at the answer
position need not preserve ranking margins, later trajectories, or task
accuracy. The Qwen intervention population is reused from prior evaluation,
and its confidence intervals do not support a general behavioral-recovery
claim.

The protection bound $k_{\max}=N-2F$ is geometric rather than causal. It counts
how many positions can be protected while leaving enough positions for $F$
disjoint folds. Actual fold safety depends on which positions are protected,
the adjacency rules, and the selector's legal-boundary definition. Passing this feasibility check does not define a universal safe threshold or ratio.

Oracle experiments use privileged correctness information and provide only an
upper-bound diagnostic. Their 93.2\% result at $R=1.5$ does not yield a
deployable selector. Failures of learned gating, pairwise structure, localized
patching, and the tested addressability intervention reject those
operationalizations without identifying the full mechanism.

The matched comparisons use one model-specific compressor per checkpoint. No
complete layer sweep, ratio sweep, multiple compressor-training-seed study, or
common large-scale free-running protocol was performed. Pythia recompresses the
evolving sequence during generation, while the cached Qwen and SmolLM2 path
compresses the prompt and appends generated states as singletons, preventing a
protocol-matched cross-family generation claim.

The Qwen3-1.7B training study has three independently trained compressors per
ratio, but the other checkpoint studies use one compressor-training seed each;
their endpoint intervals therefore do not include compressor-fit uncertainty.
Moreover, the MLP-only control shows that a generic learned state transform can
improve fixed-target likelihood without sequence shortening. The Full-versus-
Native NLL contrast must therefore not be interpreted as a pure compression
effect.

Systems results depend on the Python implementation, one RTX 4090, BF16,
context length, and fixed 16-token outputs. Qwen3-8B 8K timing was stopped by a
predefined efficiency rule, and its 2K E2E comparison used unequal output
lengths. There is no fused production kernel or production-grade comparison
with other pruning, merging, or compression systems. Structural KV reduction
is established; universal latency benefit, a larger usable context window,
production readiness, and systems state of the art are not.

The reported negative studies and endpoint disagreements do not establish a
complete causal account. The work defines no universal preservation threshold,
safe compression ratio, scaling rule, or semantic-compression optimum. The
minimum sufficient latent sequence length may depend jointly on model, task,
prompt, layer, and endpoint.

\section{Conclusion}
\label{sec:conclusion}

SemanticFold demonstrates that external tokenization and deeper computational
granularity can be separated, but also that preservation has no single scalar
certificate. Across the tested checkpoints, fixed-target likelihood,
answer-distribution proximity, linear accessibility, reasoning accuracy, and
free-running behavior can move differently. The NLL decomposition attributes
much of Qwen's favorable fixed-target result to residual adaptation rather than
shortening alone. Tail protection reduces output-distribution divergence
without a reliable behavioral recovery, and matched-ratio controls do not show
a reliable advantage for cosine-selected boundaries.

The evidence supports exact-budget latent shortening, structural upper-layer
KV savings, and bounded local runtime gains. It does not supply a universal
safe ratio, scaling law, causal mechanism, or production state-of-the-art
claim. The practical requirement is checkpoint-specific validation at the
ratio and endpoint that matter. The scientific question that remains is which
latent sequence granularity a particular model, task, and functional endpoint
actually requires.

\section*{Code Availability}

The SemanticFold implementation includes the compression runtime, evaluation
scripts, frozen compressor checkpoints, and reproducibility utilities used in
this study. These materials are being prepared for public release. No
repository URL or archival identifier is asserted in this version.

\balance
\bibliographystyle{plain}
\bibliography{source/references}
\clearpage
\onecolumn
\appendix
\section{Formal Folding Procedure}
\label{app:fold}

For path edges $e_i=(i,i+1)$ with score $s_i$, define $D[i,f,b]$ as the
maximum score over the first $i$ states using $f$ folds, where $b$ records
whether state $i$ is already matched. Standard skip/take transitions enforce
non-overlap. A solution is admissible only when it realizes the integer budget
$F=N-M$. When no positive fold budget is legal, $F=0$ is a valid outcome;
feasibility does not require every sequence to exercise the compressor.

\section{Populations and Rate Semantics}
\label{app:population}

\begin{table}[h]
\centering
\caption{Scientific evaluation populations.}
\small
\begin{tabular}{lrrrl}
\toprule
Study & Prompts & Seeds & Rates & Bootstrap unit\\
\midrule
Qwen reasoning rate sweep & 100 & 3 & 10 & prompt cluster\\
Qwen LM rate sweep & 50 batches & 3 & 10 & fixed batch\\
Finite-label interventions & 500 & 1 & 7 & stratified prompt\\
SmolLM2 reasoning rate sweep & 300 & 1 & 6 & prompt\\
SmolLM2 LM rate sweep & 60 records & 1 & 6 & record\\
Fixed continuation & 50/48 & 3/1 & 4/3 & record\\
Answer probe & 300/240 & 1 & 2/3 & task-stratified prompt\\
Original random boundary & 500/300 & 5 & 1/model & task-stratified prompt\\
Matched-ratio boundary & 300/model & 1 & 2/model & task-stratified prompt\\
Qwen3-8B fixed/reasoning & 50/300 & 1 & 4 & record/prompt\\
Qwen3 matched probe & 300 & 1 & 2/model & task-stratified prompt\\
Pythia matched probe & 300 & 1 & 2/model & task-stratified prompt\\
Pythia fixed/reasoning & 50/300 per model & 1 & 2/model & record/prompt\\
\bottomrule
\end{tabular}
\end{table}

Requested and effective ratios are distinct. The exact fold budget, integer
sequence lengths, and matching constraints induce small differences between
$R$ and $\Reff$. Machine-readable evaluation outputs retain both fields where
applicable.

\section{Compressor Architecture and Training Protocols}
\label{app:compressor-training}

All compressors instantiate the shared map in
Section~\ref{sec:compressor-training} with inner width 1,024, BF16 execution,
and a frozen backbone. The uncompressed backbone is the teacher on the same
examples and continuation tokens. Cross-entropy is evaluated on continuation
tokens; KL matches teacher and compressed-student continuation logits; cosine
loss matches their final hidden states. No early stopping is used. Candidate
checkpoints are chosen only by the stated validation criterion.

\begin{table*}[h]
\centering
\caption{Checkpoint-specific compressor training. CE, KL, and HC denote
continuation cross-entropy, teacher KL, and hidden cosine loss.}
\label{tab:compressor-training}
\scriptsize
\resizebox{\textwidth}{!}{%
\begin{tabular}{p{0.115\textwidth}p{0.14\textwidth}p{0.095\textwidth}p{0.12\textwidth}p{0.11\textwidth}p{0.17\textwidth}p{0.18\textwidth}}
\toprule
Checkpoint & Frozen data split & Lengths; train rate & Seeds & Objective & Optimizer and budget & Boundary and selection rule\\
\midrule
Qwen3-1.7B & WikiText-103 streaming train; WikiText-103 validation & 256; 192/64; $R\in\{1.5,1.75,2.0\}$ & 101, 202, 303 independently at each rate & CE $+1.0$ KL, $T=1$, $+0.1$ HC & AdamW, $10^{-4}$, weight decay .01, 20-step warmup then cosine decay, clip 1, batch 1, 1,000 steps & Deterministic contiguous training groups; minimum validation PPL among steps 20, 50, 100, 200, 400, 600, 800, 1,000; test unused\\
SmolLM2-1.7B & Frozen WikiText-derived manifest: 512 train and 128 validation records & 128; 96/32; $R=1.5$ & 20261002 & CE $+0.5$ KL, $T=2$, $+0.1$ HC & AdamW, $3\!\times\!10^{-4}$, weight decay .01, no scheduler, clip 1, batch 1, 150 steps & Cosine exact-budget folds; minimum total validation loss at steps 30, 75, or 150; step-30 nondegeneracy gate only\\
Qwen3-8B & WikiText-103 streaming train; 20 frozen WikiText validation blocks & 256; 192/64; $R=1.5$ & 20261003 & CE $+1.0$ KL, $T=1$, $+0.1$ HC & AdamW, $10^{-4}$, weight decay .01, 20-step warmup then cosine decay, clip 1, batch 1, 100 steps & Deterministic contiguous training groups; minimum mean validation NLL at steps 20, 50, or 100\\
Pythia-1.4B / 6.9B & Frozen WikiText text pool; consecutive nonoverlapping train and validation blocks & 256; 192/64; $R=1.5$ & 20261005, separately per checkpoint & CE $+1.0$ KL, $T=2$, $+0.1$ HC & AdamW, $10^{-4}$, weight decay .01, 20-step warmup then cosine decay, clip 1, batch 1, 100 steps & Cosine exact-budget folds; minimum mean continuation NLL on the first 20 validation blocks at steps 20, 50, or 100\\
\bottomrule
\end{tabular}}
\end{table*}

Qwen3-1.7B selected steps 1,000, 800, and 1,000 for the three $R=1.5$
seeds. SmolLM2, Qwen3-8B, Pythia-1.4B, and Pythia-6.9B each selected the final
candidate checkpoint. Gradient accumulation is one throughout. The
Qwen3-1.7B, Qwen3-8B, and Pythia runs use cosine learning-rate decay after
warmup; SmolLM2 uses a constant learning rate. The resulting trainable counts
are 4,200,448 at hidden size 2,048 and 8,400,896 at hidden size 4,096.

\section{Statistical Definitions}
\label{app:statistics}

For prompt $i$, let $y_i(R)\in\{0,1\}$ denote correctness. The paired accuracy
difference is
\[
\widehat{\Delta}(R)=n^{-1}\sum_i[y_i(R)-y_i(1)].
\]
Confidence intervals are percentile intervals from 10,000 prompt-level
bootstrap draws, clustered by prompt when compressor seeds repeat the same
item. Regression and recovery counts are the off-diagonal cells
$y_i(1)=1,y_i(R)=0$ and $y_i(1)=0,y_i(R)=1$. McNemar tests use those cells;
families of rate comparisons use Benjamini--Hochberg correction. Equivalence
bands are pre-specified composite gates and are never inferred from a
non-significant difference alone.

\section{Complete Qwen Endpoint Curves}
\label{app:qwen}

\begin{table}[h]
\centering
\caption{Complete Qwen3-1.7B endpoint curves.}
\small
\begin{tabular}{rrrrr}
\toprule
$R$ & Accuracy & $\Delta$ pp & PPL & AUC\\
\midrule
1.0&.410&0.00&14.560&.995\\
1.1&.400&-1.00&11.728&.977\\
1.2&.417&+0.67&11.814&.970\\
1.3&.387&-2.33&11.972&.967\\
1.4&.373&-3.67&12.174&.964\\
1.5&.357&-5.33&12.385&.966\\
1.6&.327&-8.33&12.522&.951\\
1.7&.303&-10.67&12.717&.964\\
1.8&.297&-11.33&12.803&.966\\
1.9&.313&-9.67&12.965&.969\\
\bottomrule
\end{tabular}
\end{table}

The $R=1.5$ accuracy-difference interval is $[-15.33,4.33]$ percentage
points. Materiality also uses the pre-specified margin and regression
criteria; the classification does not rest on a conventional
null-hypothesis rejection alone.

\section{Cache Geometry and Systems Protocol}
\label{app:systems-protocol}

For generation step $t$, expected layer-wise cache lengths are
\[
L_\ell(t)=
\begin{cases}
N+t, & \ell<K,\\
M+t, & \ell\ge K.
\end{cases}
\]
Independent verification reconstructed every layer length. At SmolLM2
$R=1.2$, measured KV ratios were 0.88895 at 2K and 0.88378 at 8K. The
corresponding E2E ratios were 0.88568 and 0.87973.

The direct Native comparison uses BF16 on one RTX 4090 at $R=1.3$. Both
arms receive the same repeated fixed token sequence and generate exactly
16 greedy tokens; EOS does not shorten a run. At each of 2K and 8K, one
warmup precedes five retained alternating-order measurements. KV bytes are
measured from populated BF16 caches. Cached-versus-stateless fidelity uses
the same checkpoint and compressor and compares first-token agreement,
16-token trajectories, task accuracy, and whole-trajectory exactness.

\section{Bounded Generation Protocol}
\label{app:free-running}

The SmolLM2 diagnostic uses 100 manually constructed prompts, with ten prompts
in each of ten categories: short, factual, continuation, simple reasoning,
multi-step reasoning, instruction, code, Unicode, medium, and long. It uses
\texttt{HuggingFaceTB/SmolLM2-1.7B} at revision
\texttt{effd688a12921b4cc83e3312b6feb579f70f9c71}, insertion layer 7,
BF16, and SDPA. Tokenization uses \texttt{GPT2TokenizerFast}, no chat
template, and \texttt{add\_special\_tokens=false}. Decoding uses
\texttt{torch.argmax}, no sampling, temperature zero, seed 20261002, and
16 generated tokens.

\begin{table}[h]
\centering
\caption{SmolLM2 $R=1.3$ free-running diagnostic counts.}
\small
\begin{tabular}{lr}
\toprule
Classification & Count\\
\midrule
Normal text & 93\\
Immediate EOS & 2\\
Token loop & 2\\
Printable-garble diagnostic & 3\\
\bottomrule
\end{tabular}
\end{table}

Cached/stateless first-token agreement is 100/100 and full 16-token trajectory
agreement is 85/100. No run contains NaN or Inf. The printable-garble rule is
deterministic but heuristic, and these counts must not be interpreted as a
universal failure rate.

\section{Cross-Checkpoint Protocol}
\label{app:large-model}

\paragraph{Qwen3-8B.}
The Base checkpoint revision is
\texttt{49e3418fbbbca6ecbdf9608b4d22e5a407081db4}. It uses $K=8$ of 36
layers, BF16, SDPA, and one checkpoint-specific compressor. An
outcome-independent implementation correction equalized the hidden-state
split and final-projection contracts before confirmatory scoring. The $R=1$
gate is bitwise exact.

\paragraph{Matched Qwen3 comparison.}
The exact 300-prompt population, prompt-ID folds, task-local classes, and
$R=1$/$1.3$ protocol are shared across Qwen3-1.7B and Qwen3-8B.
The 10,000-draw paired interaction analysis uses seed 20261003.

\paragraph{Independent Pythia family comparison.}
Pythia-1.4B revision
\texttt{fedc38a16eea3bd36a96b906d78d11d2ce18ed79} and Pythia-6.9B revision
\texttt{c0e3eee36dc47af0c49f361c74cfe459c09f7f23} use $K=5$ and $K=7$,
respectively. Training, bootstrap, and permutation seed is 20261005; there
are 10,000 bootstrap draws and 100 permutations per scale. Outcome-independent
corrections to generation projection and analysis adaptation are applied
before confirmatory scoring.

All three studies use one RTX 4090. No family pooling, parameter-count
regression, additional ratio, or layer sweep is performed.

\section{Selected Internal-Granularity Methods}
\label{app:related-comparison}

\begin{table*}[h]
\centering
\caption{Selected representative sequence- and state-reduction methods with
directly verifiable intervention descriptions. The table is illustrative,
not comprehensive, and is not a performance ranking.}
\label{tab:related-comparison}
\scriptsize
\resizebox{\textwidth}{!}{%
\begin{tabular}{lllll}
\toprule
Method & Input unit & Internal action & Backbone/training & Primary objective\\
\midrule
MrT5 & bytes & learned encoder-state deletion & continued pretraining &
byte-model efficiency and tasks\\
Dodo & tokens & dynamic hidden-state compression & parameter-efficient
adaptation & context compression and tasks\\
LazyLLM & tokens & dynamic prompt-state pruning & inference method &
long-context acceleration\\
SlimInfer & tokens & layer-wise hidden-state pruning & inference framework &
TTFT/E2E and memory\\
AutoCompressors & tokens & continuous summary vectors & compressor training &
latent context memory\\
ByteFlow & bytes & adaptive coding-rate chunking & trained hierarchy &
tokenizer-free language modeling\\
H2O & tokens & KV-entry eviction & inference policy & bounded decode cache\\
SemanticFold & tokens & exact-budget adjacent-state replacement & frozen
backbone; small compressor & controlled functional characterization\\
\bottomrule
\end{tabular}
}
\end{table*}

Internal sequence shortening, variable hidden-state counts, and adaptive
computational granularity are established prior art. SemanticFold's narrower
contribution is a frozen-backbone, exact-budget intervention evaluated jointly
with a true $R=1$ oracle, identical-target continuation, task-local cross-fitted
probes, reasoning, bounded generation, matched random boundaries, matched
checkpoint comparisons, and an independent second family.

\section{Intervention and Decomposition Protocol}
\label{app:interventions}

The finite-label intervention population contains 450 prompts from nine
50-example BBH tasks. A split-model identity gate compares native logits with
the original model; all 1,100 model--prompt checks pass. The full SemanticFold
and pair-only arms reuse the same lower-layer states and selected boundaries.
Recorded quantities include choice scores, correctness, margins, KL, token
counts, fold boundaries, singleton counts, and protected-tail lengths.

The task-stratified bootstrap resamples within task using 10,000 draws and
seed 20261004. Exact two-sided McNemar tests compare pair-only with the full
SemanticFold arm, with Holm correction across four primary comparisons. All
four adjusted values equal one. The Qwen tail arm has 64 wins and 57 losses
against full SemanticFold, while the SmolLM2 tail arm has 31 and 37. The
pair-only arms have 18/22 and 22/21, respectively.

The fixed-continuation decomposition uses 50 Qwen records (192 prefix and
64 continuation tokens, $R=1.7$) and 48 SmolLM2 records (96 prefix and
32 continuation tokens, $R=1.2$). Native, full SemanticFold, pair-only,
MLP-only, and mean-merge arms share each record's prefix and continuation
IDs. MLP-only transforms all prefix states without shortening. Pair-only
restores every singleton from its lower-layer source. Mean merge averages
the two lower states at the selected boundaries, copies singletons, and keeps
the same last-source position policy.

Record-level mean-NLL contrasts use 10,000 percentile-bootstrap draws and seed
20261004. Full SemanticFold minus MLP-only is $+0.08161$ for Qwen (95\% CI
$[+0.05914,+0.10721]$) and $+0.07699$ for SmolLM2
($[+0.03003,+0.14393]$). Mean merge minus Native is $+0.10124$
($[+0.07635,+0.12801]$) and $+0.27145$ ($[+0.09549,+0.50145]$),
respectively.

\section{Distribution--Behavior Analysis}
\label{app:distribution-behavior}

The distributional analysis uses the exact 450-prompt Qwen3-1.7B finite-label
population and answer-position logits at $R=1.7$. For every prompt and arm,
it computes
$D_{\mathrm{KL}}(p_{\mathrm{Native}}\Vert p_{\mathrm{arm}})$ over the full
vocabulary. Mean and median intervals use 10,000 prompt bootstrap draws with
seed 20261005. Accuracy contrasts use the same task-stratified paired
procedure.

Tail protection reduces mean KL from 1.404 to 0.527 and median KL from
1.238 to 0.497, while its accuracy change from full SemanticFold is
$+0.0156$ with 95\% CI $[-0.0311,0.0622]$. MLP-only has mean KL 0.097 and
accuracy .3489. Distributional proximity is informative under some
interventions but is not a sufficient certificate for behavioral recovery.

\section{Matched-Ratio Boundary Protocol}
\label{app:matched-r}

The matched control uses the same first 300 eligible prompts for every arm and
checkpoint. It evaluates Qwen3-1.7B and SmolLM2-1.7B at $R=1.3$ and $R=1.7$.
A deterministic uniform legal matching is sampled once per prompt with seed
20261005. The random and cosine arms share the checkpoint, compressor,
readout, continuation target, $N$, $M$, and $F$. Split Native logits are
checked bitwise against the full model. The task-stratified paired bootstrap
uses 10,000 draws.

Cosine-minus-random accuracy contrasts are $+0.0200$
($[-0.0367,0.0767]$) and $+0.0233$ ($[-0.0400,0.0867]$) for Qwen at
$R=1.3$ and $1.7$, and $+0.0233$ ($[-0.0400,0.0867]$) and $-0.0100$
($[-0.0667,0.0467]$) for SmolLM2.

\section{Mechanism-Directed Intervention Protocols}
\label{app:mechanism-protocols}

This section gives the operational protocol behind every proposition in
Table~\ref{tab:mechanisms}. Unless stated otherwise, these studies use the
frozen Qwen3-1.7B checkpoint at insertion layer 6, BF16, compressor seeds 101,
202, and 303, and finite-label accuracy or correct-answer margin. They diagnose
the tested operationalization only.

\paragraph{Risky-fold diagnostic.}
The population is the fixed 100-prompt reasoning suite at $R=1.5$. Within
native-correct/compressed-incorrect cases, single-group restoration supplies a
group-level change in answer margin. The positive label is the top decile of
restoration benefit within each prompt--seed case. Features include group
width, compressor-weight entropy and maximum weight, mean pairwise cosine,
hidden and norm variance, position, lexical/token structure, and prompt-level
context. A standardized class-balanced logistic regression with random state
zero is fitted on a deterministic category-stratified 60/20/20 prompt split.
Held-out ROC AUC is 0.645. The result is a restoration association, not a
causal or deployable selector.

\paragraph{Online rate controller.}
For requested rates $R\in\{1.2,1.3,1.4,1.5\}$, the controller computes the
exact fold budget $\operatorname{round}_{\mathrm{even}}(N-N/R)$ and ranks
adjacent cosine edges, breaking ties by source index, subject to non-overlap.
The audit spans the frozen reasoning suite, 50 WikiText language-model blocks,
and repeated 1K, 2K, 4K, and 8K system prompts. There is no fitted model or
outcome label. The endpoint is absolute requested-versus-realized rate error:
the overall mean falls from 0.2079 to 0.0020, and the largest
domain--target mean is 0.0057.

\paragraph{Learned gate.}
After two technical exclusions, 39 native-correct prompts are split by prompt
into 24/7/8 train/validation/test items; every legal boundary and all three
compressor seeds inherit the same split. Candidate families comprise geometry,
variance, prompt-relative, Q/K/V, token-structure, and compressor-diagnostic
features. Validation selects a logistic regression with $C=10$ using all cheap
features except attention features; the target is the upper-quartile harm
indicator. On the eight test prompts at $R=1.5$, learned, cosine, random, and
oracle accuracies are 70.83\%, 66.67\%, 62.08\%, and 100\%. Learned-minus-
cosine changes by seed are $+12.5$, $-25$, and $+25$ percentage points, so the
pre-specified stability criterion fails.

\paragraph{Oracle-safe boundaries.}
The same 39-prompt population contains 5,190 legal boundaries and 15,570
seed-specific single-fold interventions. Boundary harm is the native
correct-answer margin minus the margin after that single fold. An exact
cardinality dynamic program minimizes measured harm at
$R\in\{1.2,1.3,1.4,1.5\}$, followed by an actual multi-fold model execution.
At $R=1.5$ the oracle accuracy is 0.932; a 5,000-draw prompt-cluster bootstrap
quantifies uncertainty. Because correct-answer margins define the selector,
this is privileged-label recovery headroom.

\paragraph{Stable pair structure.}
The pair study uses the same 39 prompts and three seeds. It records 178,227
seed-specific pair interventions from 8,347,680 possible cases, with complete
graphs on 12 prompts and coverage of all prompts. Pair interaction is the
departure of a two-fold margin effect from the sum of its single-fold effects.
Cross-seed macro Spearman correlation is 0.1403 and sign agreement is 0.5175;
pairwise reconstruction is worse than the additive baseline. At $R=1.5$ the
pairwise oracle reaches 0.6111 versus 0.9167 for the measured single-edge/full
oracle. The tested pair structure is therefore seed-specific rather than
stable.

\paragraph{Single-fold adapter.}
The 24/7/8 prompt split is retained. A matched representation table contains
2,404 train/validation boundaries and 7,212 seed rows. Low-rank residual maps
of ranks 4, 16, and 64 predict the answer-independent natural pair-mean
upper-input representation; validation normalized hidden MSE selects rank 64
(264,192 parameters), with training seed 271828 plus the rank. On the eight
test prompts, a 31415926-seed prompt bootstrap gives accuracy change $-0.0068$
with 95\% CI $[-0.0218,0.0068]$ and a 22.21\% perplexity increase. Only
single-fold interventions are tested; the positive-effect subset has four
prompts, and no multi-fold claim is made.

\paragraph{Localized patch.}
One hundred prompts are deterministically divided within category into 40
train, 20 validation, and 40 sealed test items. Source--donor seed pairs are
$303\leftarrow202$ and $202\leftarrow101$ at rates 1.15, 1.5, and 1.75.
Candidate patches cover Q, K, V, attention output, MLP output, residual input,
and block output over upper layers 6--27; attention logits and probabilities
are unsupported by the instrumentation. Train/validation screening includes
self-donor, shuffled-donor, and random-layer controls and requires positive
recovery at at least two rates. No candidate is eligible, so the sealed test
set remains unopened.

\paragraph{Addressability intervention.}
The same 40/20/40 split and three seeds are used at
$R\in\{1,1.15,1.5,1.75\}$, but only the 60 train/validation prompts are opened.
A deterministic, answer-independent boundary score sums six unit-weight
crossings: distinct object, role type, relation endpoint, operator argument,
clause, and reference. Exact-budget dynamic programs either minimize
(preserve) or maximize (destroy) this score; random and cosine policies are
controls. With 10,000 prompt bootstrap draws and seed 2718281, the
preserve-minus-destroy accuracy effect is $-0.025926$ with 95\% CI
$[-0.103704,0.057407]$. This rejects the tested annotation proxy on the opened
split and does not generalize to all forms of addressability.

\paragraph{Structural-feature prediction.}
On the same 60 opened prompts, the target is reasoning-margin loss relative to
$R=1$. Standardized ridge regression with $\alpha=1$ uses deterministic
prompt-grouped five-fold cross-fitting. M1 contains effective rate and semantic
similarity; M2 adds addressability destruction; M3 adds change in crossing
degrees of freedom and the six category terms. Added structure is considered
reproducible only if out-of-fold $R^2$ improves by at least 0.02 and the lower
bound of the prompt-bootstrap MAE-improvement interval exceeds zero. M2 raises
$R^2$ by 0.0344 but its MAE interval crosses zero; M3 changes $R^2$ by
$-0.01046$. Neither model passes the joint criterion.

\section{Budget Rounding and Protection}
\label{app:budget-protection}

The implementation computes the fold count first using Python half-to-even
rounding, then derives the compressed length. Half-integer tests verify
$F=2,4,4,6$ for $N=5,7,9,11$ at $R=2$. The 450 Qwen tail records at
$R=1.7$ have $k_{\max}$ min/median/max of 7/27/48; the requested tail length
16 is realized as 7--16. The 450 SmolLM2 records at $R=1.3$ have
$k_{\max}$ 25/89/152 and realize all 16 positions. Every record passes the
declared fold-safety predicate.

\end{document}